\def\year{2020}\relax
\documentclass[letterpaper]{article} 
\usepackage{aaai20}  
\usepackage{times}  
\usepackage{helvet} 
\usepackage{courier}  
\usepackage[hyphens]{url}  
\usepackage{graphicx} 
\usepackage{graphicx}  
\usepackage{ragged2e}
\usepackage{enumitem}
\usepackage{color}
\definecolor{guzcolor}{rgb}{0.0,0.2,0.8}
\definecolor{nazcolor}{rgb}{0.8,0.0,0.2}

\usepackage[utf8]{inputenc}
\usepackage{graphicx}
\usepackage[justification=centering]{caption}
\title{Entity Embedding as Game Representation}

\author{Nazanin Yousefzadeh Khameneh and Matthew Guzdial\\
Department of Computing Science, Alberta Machine Intelligence Institute (Amii) \\
University of Alberta, Canada\\ \{nyousefz, guzdial\}@ualberta.ca \\}
 
\begin{document}

\maketitle
\begin{abstract}

Procedural content generation via machine learning (PCGML) has shown success at producing new video game content with machine learning.
However, the majority of the work has focused on the production of static game content, including game levels and visual elements. 
There has been much less work on dynamic game content, such as game mechanics.
One reason for this is the lack of a consistent representation for dynamic game content, which is key for a number of statistical machine learning approaches.
We present an autoencoder for deriving what we call ``entity embeddings'', a consistent way to represent different dynamic entities across multiple games in the same representation.
In this paper we introduce the learned representation, along with some evidence towards its quality and future utility.

\end{abstract}

\section{Introduction}

Generating game content using Machine Learning (ML) models trained on existing data is referred to as Procedural Content Generation via Machine Learning (PCGML). 
PCGML has shown success in both game development and technical games research \cite{karth2017wavefunctioncollapse,summerville2018procedural}. 
Most of the prior work has focused on level generation and not mechanic generation or entire game generation.
One reason is that there are few available shared representation frameworks across games. 
Games vary significantly from one another, across multiple game systems, game genres, and particular instances of a genre. 
We also cannot use the mechanics of just one individual game, as modern machine learning approaches rely on training data sizes that far surpass what a single game could provide.
Given all of this, it is difficult to represent functional pieces of different games in the same representation.
A suitable shared data representation for different games would make it possible to do different PCGML-related tasks more easily like novel game generation, novel game mechanic generation, transferring knowledge between games, automated reasoning over games, and so on. \par

There is not a large body of prior work in data representation for dynamic game content in PCGML.
Guzdial and Riedl presented a method in which a new graph-like game representation is learned in order to generate new games by recombining existing ones \cite{guzdial2018automated}. 
However, the graph-like representation is not well-suited to statistical machine learning methods.
Osborn et al.  proposed a system to derive a complete model of game structure and rules as a new representation for games \cite{osborn2017automated}. 
This approach has only been proposed, not yet implemented. 
Recently, there have been some projects on game generation with PCGML using the Video Game Description Language (VGDL) \cite{machado2019pitako}, however we note that all of the dynamic knowledge had to be hand-authored, rather than learned from existing games.

Ideally, we would like to have a representation that would allow us to represent machine-learned knowledge of dynamic game elements that is suitable to statistical machine learning tasks.
In this paper we aim to learn such a representation of dynamic game entities as low dimensional vectors in which mechanical information is preserved. 
We call this approach \emph{entity embedding}.
This entity embedding attempts to obtain the functional (mechanics) similarities between entities not the aesthetic (appearance) ones. 

In this paper, we use a Variational AutoEncoder (VAE) to re-represent Atari Games with an entity embedding in a lower dimension representation. 
We evaluate our approach with some similarity measures in comparison to a K-nearest neighbors-inspired baseline. 
In addition, we demonstrate some qualitative examples of the potential applications of this representation.

\section{Related Work}

In this section we focus on related PCGML approaches, and related deep neural network (DNN) based approaches for modeling the dynamic elements of games. 

In the introduction we identified that the majority of PCGML approaches have been applied to level generation or the generation of other static content. 
However, there are some prior examples that touched on the generation of dynamic game entities. 
Guzdial et al. \cite{guzdial2017game} introduced an approach to learning the rules of a game from gameplay video and then applied these learned rules to rule generation \cite{guzdial2018automated}.
Similarly, Summerville et al. employed a causal learning framework to learn certain semantic properties of various game entities \cite{summerville2017does}. They later proposed applying this as part of a pipeline to generate new games \cite{osborn2017automated}.
Recently, Bentley and Osborn \cite{bentley2019videogame} presented a corpus of such semantic game properties, which could be employed in PCGML. 
However, this dynamic information would have to be hand-authored and then applied to a PCGML problem as in the work of Machado et al. \cite{machado2019pitako}.
Comparatively, we seek to re-represent learned dynamic game information in a smaller, latent representation which can then be used for PCGML tasks.

In this work we rely on a Variational AutoEncoder (VAE). VAE's have been applied to many other PCGML level generation tasks.
They have been used to generate levels for Super Mario Bros. \cite{jain2016autoencoders,guzdial2018explainable} and Lode Runner \cite{thakkar2019autoencoder}.
Sarkar et al. employed a VAE to learn two different game level distributions and then generate content from in-between these learned distributions \cite{sarkar2020controllable}.
We also rely on learning content from multiple games, however our content is a representation of dynamic game entities instead of structural information.

Outside of PCGML, there exists work in learning to model dynamic elements of games to help automated game playing tasks.
Ha and Schmidhuber presented their approach ``World Models'' that used a VAE as part of its pipeline to learn a forward model, a way of predicting what will happen next according to the mechanics of a world \cite{ha2018world}.
These ``World Models'' were helpful in improving agent performance on playing the modelled games, but the learned representation of dynamic elements was far too large to use as input in a PCGML process.
Similarly, Go-Explore uses a VAE-like model for determining whether it has been to a particular state \cite{ecoffet2019go}. 
However, the goal is simply to use the latent embedding as a distance function comparing game states, not as a representation for PCGML.
Most recently, Kim et al. \cite{kim2020learning} presented an approach to use an augmented Generative Adversarial Neural Network (GAN) to model an entire game by separately modeling dynamic and static entities. 
We similarly seek to model dynamic entities with a DNN, but we focus on modeling the entities of multiple games instead of having a model trained to recreate one specific game. 
Further, just as with the World Models approach, the representation is too large to apply a PCGML approach on it.

\section{System Overview}

In this paper, we develop a method for embedding entities from multiple games in a 25-dimension latent vector.
We focus on the domain of Atari games to test this approach, as the games are relatively simple while still being more complex than hand-authored games in the VGDL, which makes hand-authoring knowledge from them non-viable. 
We identify this dynamic information automatically from these Atari games by running the rule learning algorithm introduced by Guzdial et al. \cite{guzdial2017game}.
We then collect information for each entity based on the learned ruleset by vectorizing the learned rule information for each entity. 
Finally, we train our VAE with this vectorized representation to obtain the latent space.  

Our trained VAE gives us our entity embedding as points in a learned 25-dimensional latent space. We
In this representation we can represent changes over entities as vectors (one entity at one point becoming another entity at another point), and whole games as graphs or point clouds (where each point is an entity in the game). 
We anticipate that these compact representations will make PCGML work that involves mechanical, dynamic, or semantic information far easier. 


\subsection{Ruleset Learning}

\begin{figure*}[tbh]
    \includegraphics[width=\textwidth, height = 5 cm]{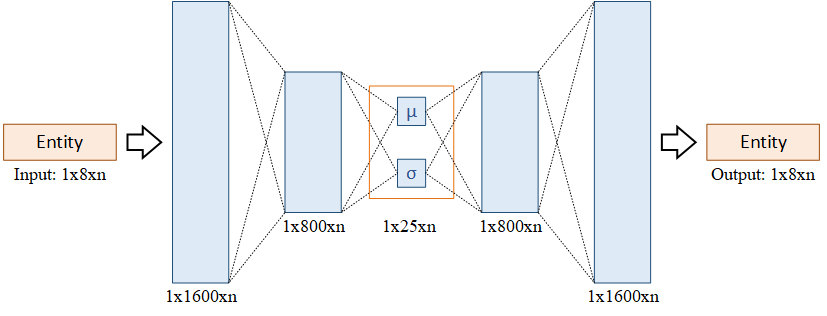}
    \caption{Architecture of our VAE}
    \label{fig:vaeArchitecture}
\end{figure*}

Since our goal is to have our representation reflect the semantics (mechanics) of the entities, we decided to obtain game rules to collect this information. Thus we make use of the game engine learning algorithm from Guzdial et al. \cite{guzdial2017game} to learn rulesets for each game. 
The algorithm tries to predict the next frame with a current engine (sequence of rules), if the predicted frame is sufficiently similar to the original one the engine remains the same, otherwise it optimizes the current engine via search. 
Each rule consists of conditional facts and effect facts.
The facts are percept-like representations that denote individual atomic units of knowledge about the game \cite{ugur2009affordance}.
For each rule to fire all the conditional facts must be true. 
Upon firing, the rule replaces one fact (the preffect) with another fact (the posteffect). 
This allows the rules to model changes in a game, like movement, entities appearing and disappearing, and changes in entity states.
that represents the mechanics of the game and from which it is possible to simulate the whole game \cite{guzdial2018automated}. 
Below is the list of the types of facts we use in this paper: 
\begin{itemize}
\item \textit{Animation} contains \textit{SizeX} \textit{SizeY} of the entity.
\item \textit{VelocityX} indicates the velocity of the entity horizontally.
\item \textit{VelocityY} indicates the velocity of the entity vertically.
\item \textit{PositionX} this fact is the value of an entity in the $x$ dimension of a frame.
\item \textit{PositionY} this fact is the value of an entity in the $y$ dimension of a frame.
\end{itemize}

For example, in the following rule (Rule X), entity A's speed in X direction will change from 0 to 5 as its conditional facts match the current game state \newline
\textit{
RULE X: \newline
VelocityXFact: [A, 0]$\rightarrow$VelocityXFact: [A, 5]\newline 
VelocityXFact: [A, 0]\newline
VelocityYFact: [A, 0]\newline
AnimationFact: [A, (8, 4, 3)]\newline
PositionXFact: [A, 79]\newline
PositionYFact: [A, 17] \newline
\newline
VelocityXFact: [B, 0]\newline
VelocityYFact: [B, 0]\newline
AnimationFact: [B, (5, 6, 3)]\newline
PositionXFact: [B, 93]\newline
PositionYFact: [B, 42] \newline
etc. \newline
}
For more information on the Engine learning process please see \cite{guzdial2017game}.

\subsection{Dataset}

In this paper, we made use of two Atari games, Centipede and Space-invaders, as represented in the Arcade Learning Environment (ALE) \cite{bellemare2013arcade}. 
We chose these two Atari games since both games have similar mechanics in which the player is a fighter who shoots at enemies.
We ran the Game Engine Algorithm on roughly 100 frames of each game to obtain the game rules. Each rule consists of some conditional facts and an effect. 
These conditional facts describe mechanical features of entities like size, velocity, position and so on.
The effect is made up of a pre-effect and post-effect, also describing mechanical features.
In frames where they are true (i.e. the mechanical features exactly match) that rule fires meaning that the post-effect replaces the pre-effect (e.g. the velocity of an entity changes).

After obtaining the game rules we ran a parser through each rule to save the mechanical information of each game entity as an integer in an individual vector of shape (1x8). 
For example, entity 'A' in game 'B' is represented as a vector which contains,\textit{ EntityID: A, SizeX, SizeY, VelocityX, VelocityY, PositionX, PositionY} and \textit{GameID: B}. 
We note that different in-game entities would generate multiple instances of this representation. 
Further, velocity and position values had to be integers as they were measured over the space of pixels.
Our goal was to represent each mechanical state that each game entity (EntityID) could legally be in according to the game rules.

During development we used two different representations of our dataset. 
First we used a one-hot encoding for GameID and EntityID while all other features remained integers. 
However we have less than 100 EntityIDs and only 2 GameIDs, we choose 100 and 10 as one-hot encoding sizes for EntityIDs and GameIDs, respectively. 
This is because of potential future studies with more entities or games. 
In the second representation we apply one-hot encoding to all features. 
All of the features are greater than zero except \textit{VelocityX} and \textit{VelocityY} which can be negative. 
We convert each entity to a vector of shape 1x1600 (8x200). 
Due to the fact that all the absolute values of the features are always less than 100, we chose 200 as one-hot encoding size (to represent positive and negative values). 
We found that the second representation results far exceeded the first for our evaluations and so we only report those.

\subsection{Training}

Autoencoders are efficient tools for dimensionality reduction.
These tools approximate a latent structure of a feature set.
We need to reduce the dimensionality of the entities in order to learn an entity representation with less variance. 
We decided to make use of a Variational AutoEncoder (VAE), as it would allow us to learn a more consistent latent space and sample novel entities from this learned distribution.
We applied VAE to our dataset to learn the parameters of a probability distribution to represent our entities. Thus it makes it possible to sample from this distribution and generate new entities. 
Since it is a generative model, we can apply this feature to PCGML tasks like generating entities similar to the input, blending the entities in the latent space and so on. 
We tried various VAE architectures for training. 
We obtained the final model empirically, which we visualize in Figure \ref{fig:vaeArchitecture}. 
As the Figure demonstrates, our architecture has one fully connected hidden layer with Relu activation in the encoder, which then feeds into a 25-dimensional embedding layer with Relu activation.
The decoder section architecture is an inverse of the encoder section, starting with a Sigmoid activation fully connected layer. 
We implemented this model with the Adam optimizer with a learning rate of 0.001 and binary cross-entropy as our loss function. 
We implemented this model in Keras. \par

\subsection{Generation}

The decoder generates 1x1600 vectors. We then use our one-hot representation for the decoder's output, by querying the generated outputs and finding the largest value in each ([[0-200][201-400][401-600][601-800][801-1000][1001-1200][1201-1400][1401-1600]]) segments and replacing it with 1 and others with 0. 
Thus, we can generate entirely novel outputs not previously seen during training.
\par

\begin{figure*}[t]
\centering
    \includegraphics[width=15cm]{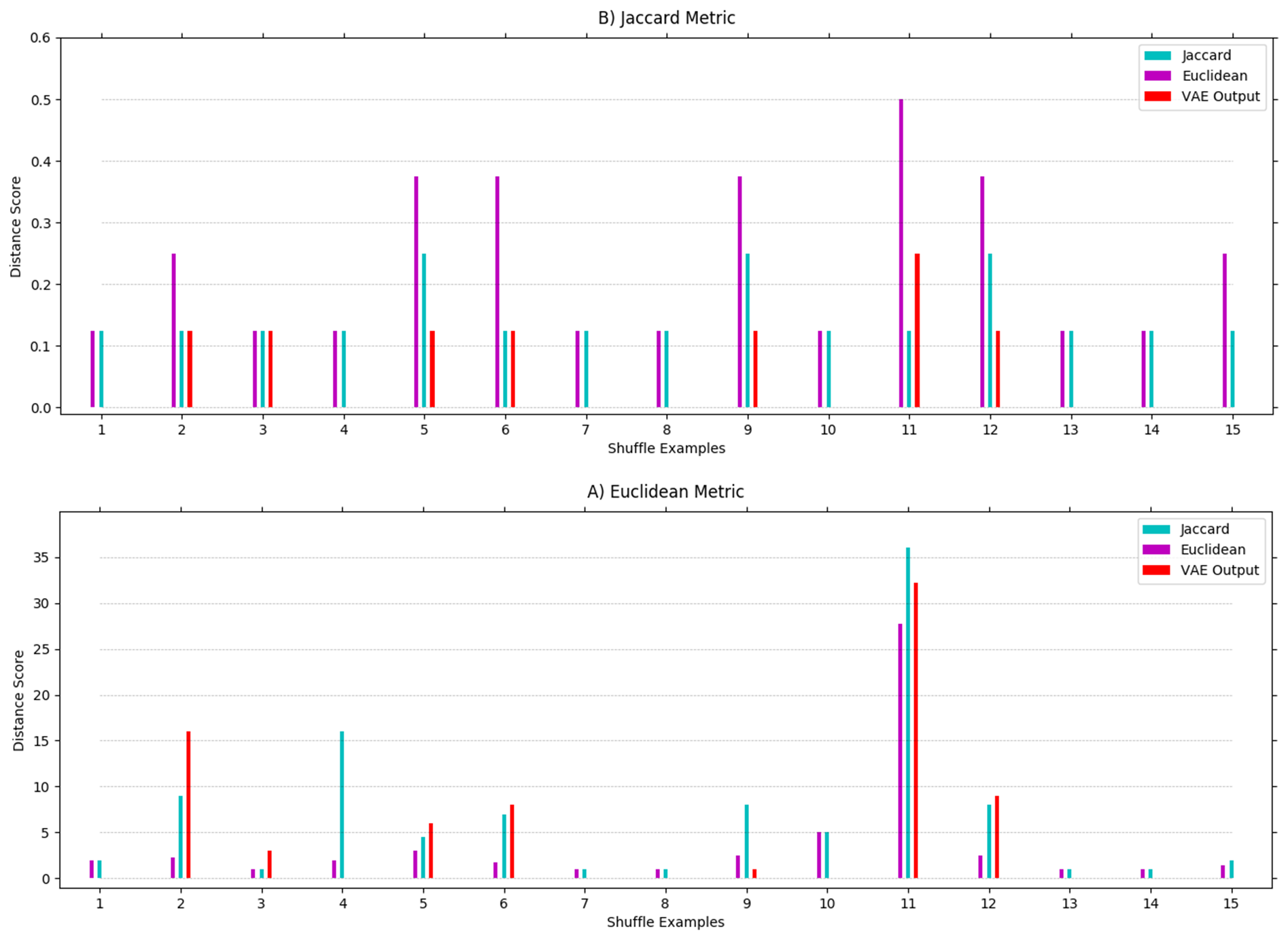}
    \caption{Visualization of 20\% of our test entities comparisons to our VAE output and our baselines.}
    \label{fig:barGraphs}
\end{figure*}

\section{Evaluation}

The entire purpose of this new entity embedding representation is to accurately represent the semantic information in a more compact representation. 
Therefore, accuracy is key.
In order to evaluate the accuracy of our VAE we ran an evaluation to compare the performance of our VAE on a held out testset of 10\% of our available data.
As a baseline, we were inspired by K-Nearest Neighbors and so selected the most similar entity from the training dataset to each entity in the testset.
To determine the most similar entity we applied two different similarity measures as below:
\begin{itemize}
\item \textit{Jaccard Similarity} that is the measure of similarity for the two sets of data based on their overlap.
\item \textit{Euclidean distance} that computes the square root of the sum of squared differences between elements of the two vectors.
\end{itemize}

We found the most similar training entity to the test entity with these two methods. 
We then compare the VAE reconstruction of the test entity and the selected training entity with the original entity.
To do this we consider (A) the number of equal values and (B) the difference of unequal values between original and predicted entity vectors. 
We employ the Euclidean and Jaccard distance functions again as comparison metrics. 
Lower values are better for both metrics as it indicates fewer differences. 
This is notably quite a strong baseline, given that many entities in the training and test sets have substantial overlap. 
We also compare the VAE with Principal Component Analysis (PCA) which is an unsupervised, non-parametric statistical technique used for dimensionality reduction in machine learning.
If the VAE is able to perform similarly or better than the closest training instance, on a test instance it has never seen before, this will indicate that the VAE has learned an accurate entity representation.


\begin{figure*}[t]
\centering
    \includegraphics[width=15cm, height =7cm]{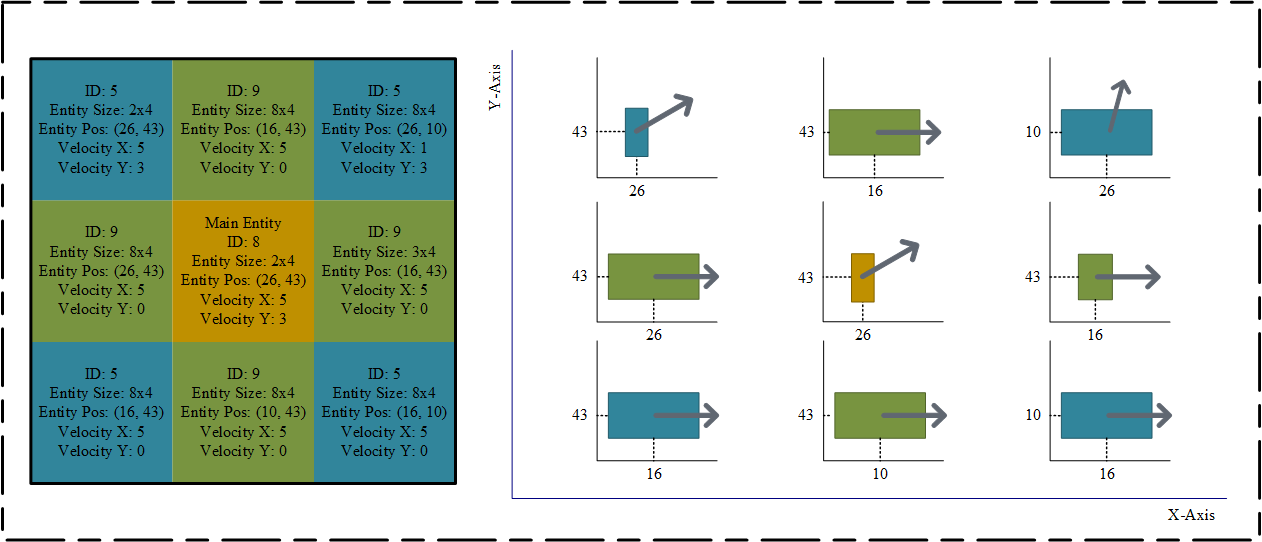}
    \caption{Eight random variations made to entity (center) in the latent space.}
    \label{fig:randomVariations}
\end{figure*}

\begin{figure}[tb]
    \includegraphics[width= 8.5 cm, height =7 cm]{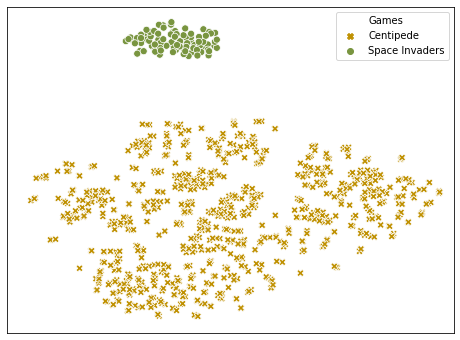}
    \caption{t-SNE Visualization of our latent space.}
    \label{fig:tSNEVisualization}
\end{figure}
\begin{figure}[tb]
    \includegraphics[width=8.5cm, height =5cm]{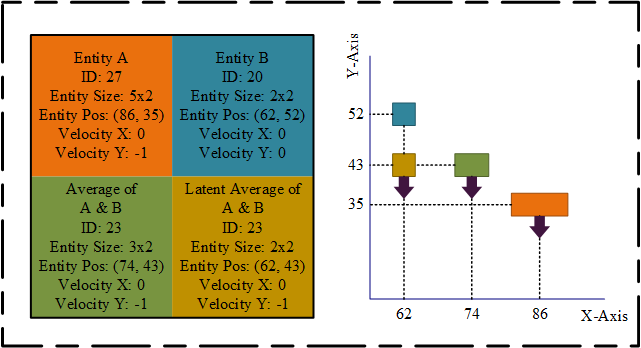}
    \caption{Comparing the average entity between two existing entities in terms of the vector representation and in the latent space. }
    \label{fig:averageEntities}
\end{figure}
\begin{figure*}[tb]
    \centering
    \includegraphics[width=16cm, height =6cm]{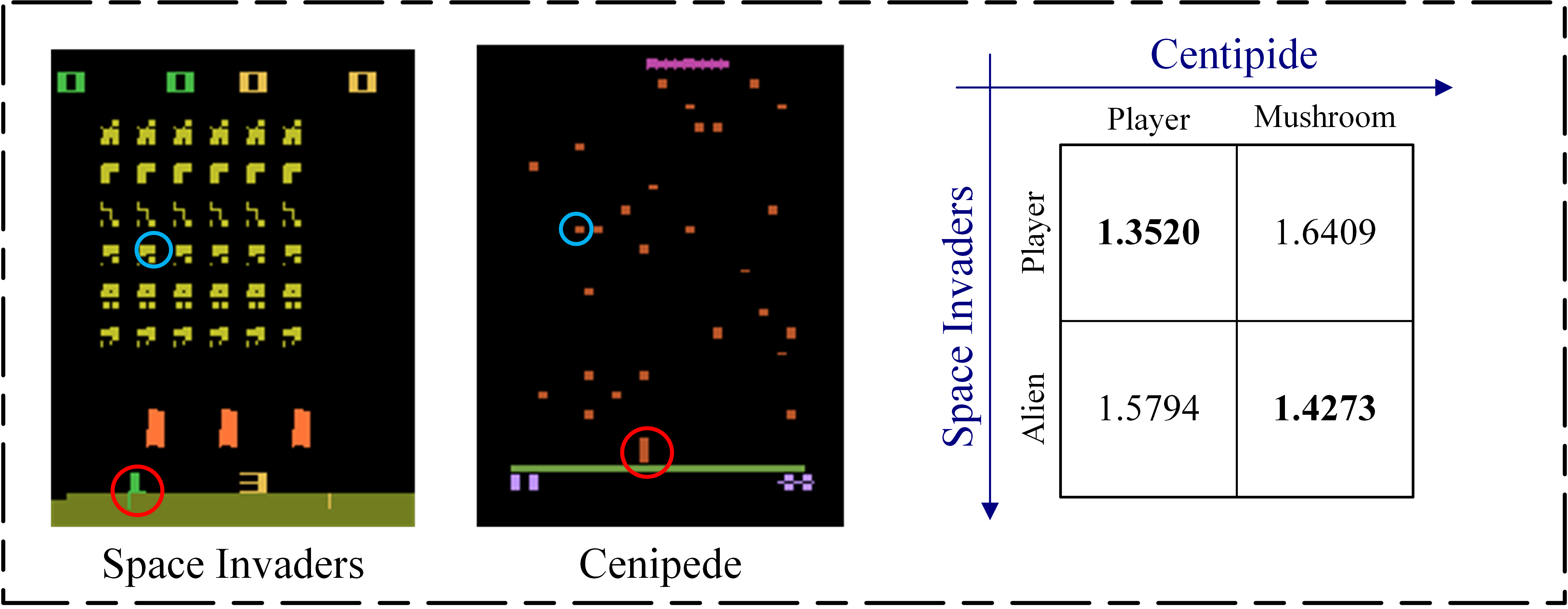}
    \caption{Two randomly selected frames of the games Centipede and Space invaders. Entities in blue circles are the Alien and Mushroom from Space invaders and Centipede respectively, both destroyable entities. Entities with red circles are player entities. The tables indicates the Euclidean distance of these entities in the latent space. }
    \label{fig:EntitySimilarity}
\end{figure*}

\begin{table}[tb]
\centering
\begin{tabular}{||c||c|c|c||} 
\hline
Metric & Jaccard Distance & Euclidean Distance \\ [0.5ex] \hline
\hline
VAE & \textbf{0.0937} & 5.6364 \\
\hline
PCA & 0.2291 & 18.0278 \\
\hline
SE Euclidean & 0.2013 & \textbf{2.8881} \\
\hline
SE Jaccard & 0.1388 & 6.0629 \\
\hline
\end{tabular}
\captionsetup{justification=centering}
\caption{Comparison of our two distance functions over our test data, comparing the output of our VAE to our baselines of the most similar training entity (SE) and PCA according to the two different distance functions.}
\label{table:1}
\end{table}


\section{Results}

The mean values across our test set for our distance scores are in Table \ref{table:1}. In the table we refer to the most similar entity in the training set to the test set as ``SE''. Thus ``SE Euclidean'' is the most similar entity in the training set to a particular test set using the Euclidean distance function. 
As is shown in the table, our VAE outperforms the other methods when we use the Jaccard distance. 
However, the SE Euclidean performs better when we use Euclidean distance, though notably we outperform the SE Jaccard baseline even with the Euclidean distance function. 
This indicates that the VAE outputs and the original entities share more equal feature values but the individual feature values at times have larger variation compared to the closest entity in the training data. 
Furthermore, our proposed VAE outperforms PCA which is another dimensionality reduction method.
We demonstrate distance score for individual randomly selected entities from our testset in Figure \ref{fig:barGraphs}. As is shown, the majority of the values of the VAE are lower compared with the baselines except a few outliers.
It is important to note that the VAE generates the exact same entity roughly 50\% of the time for the testset. \par

\section{Qualitative Examples}

In this section we display the distribution of entities in the latent space using the t-Distributed Stochastic Neighbor Embedding (t-SNE) technique. 
This technique is for dimensionality reduction and is well suited for the visualization of high-dimensional datasets \cite{maaten2008visualizing}. 
In addition, we explore the latent space by presenting some qualitative examples. \par

We provide a t-SNE visualization in Figure \ref{fig:tSNEVisualization}. 
The representation depicts the distribution of entities in the projected 2D space. 
Note that clusters correspond to the two games in our dataset, verifying the power of the model in discriminating the entities based on the GameID feature.
We also note that this indicates that we can represent games as clusters of points in this space. 
We hope to explore the possible research direction of this feature in future. \par

We also examine some qualitative examples to explore entities interpolation in the latent space. To do this, we randomly chose a pair of entities. We then calculate the average of both the vector and latent representations of the pair. As is shown in Figure \ref{fig:averageEntities}, the latent average is more like taking features from each entity while the original average is just the mean of two entity vectors (all numbers rounded down for the average). 
This indicates that our latent space is not replicating the geometric information presented in the vector representation. \par

Our second qualitative example is to analyze the surroundings of an entity inside the latent space. 
We first add various normal random vectors in the range (-0.2 to 0.2) to a randomly selected entity's embedding.
We choose this range since the entity does not change perceptively with lower ranges. 
Figure\ref{fig:randomVariations} displays the randomly selected original entity and 8 random neighbor entities around it. 
The random variations seem quite consistent in terms of sizes and IDs. 
This indicates that similarly shaped entities with similar IDs are closer together inside the latent space. This also applies to velocity and position features. \par
Our third qualitative example indicates that entities with similar mechanical features from different games are more similar in the latent space compared to entities with different mechanical features from the same game. 
This indicates that our latent space places mechanically similar entities closer to one another in its learned latent space.



\section{Future Work}

We trained a VAE on mechanical features of entities in order to derive an entity embedding.
We argue that this entity embedding can be a shared representation that enables various PCGML-related tasks. We list potential directions for future work below. \par
\begin{itemize}
\item \textit{Entity Blending:} We can generate new entities based on existing ones as is shown in the Qualitative Examples section. We plan to run another study to analyze if we can use generated entities to generate new types of rules or levels of an existing game, or entirely new games.
\item \textit{Transfer Learning}: As we discussed in the Ruleset section, the embedding is based on mechanical features of entities in each rule of a game. 
Each rule references a set of entities (group of conditional facts) together in a frame which causes an effect. We might expect similar mechanical effects if we have entities from another different game with a similar latent representation. We anticipate a need for another study to investigate this.
\item \textit{Extending the Dataset:} We trained this model on around 100 frames of two Atari games with similar mechanics. Extending our dataset by adding extracted rules of other similar and dissimilar games is another potential future direction.
\end{itemize}


\section{Conclusions}

In this paper we presented an approach to derive an entity embedding, a latent representation of in-game entities based on their mechanical information, via a Variational AutoEncoder (VAE). 
We discussed how we trained this VAE, and evaluated the entity embedding in terms of its accuracy at representing unseen test entities. 
We found that the VAE outperformed our K-Nearest Neighbor inspired baselines in most cases, indicating a general learned embedding.
We hope that this representation will lead to new applications of PCGML involving game mechanics.

\section*{Acknowledgements}

We acknowledge the support of the Natural Sciences and Engineering Research Council of Canada (NSERC), and the Alberta Machine Intelligence Institute (Amii). 

\bibliographystyle{aaai.bst}
\bibliography{main.bib}

\end{document}